\documentclass{article} % For LaTeX2e
\usepackage{colm2024_conference}

\usepackage{microtype}
\usepackage{xurl}
\usepackage[hidelinks]{hyperref}
\usepackage{booktabs}
\usepackage{listings}
\usepackage{lscape}
\usepackage{pdflscape}

\title{GEITje 7B Ultra: A Conversational Model for Dutch}

\author{Bram Vanroy\\
KU Leuven, Dutch Language Institute\\
Leuven, Belgium\\
{\tt \href{mailto:bram.vanroy@kuleuven.be}{bram.vanroy@kuleuven.be}} \\
}

\colmfinalcopy % Uncomment for camera-ready version, but NOT for submission.
\begin{document}

\maketitle

\begin{abstract}
Language models have rapidly evolved, predominantly focusing on English while often neglecting extensive pretraining in other languages. This approach has required initiatives to adapt powerful, English-centric models to other linguistic contexts through finetuning. For Dutch, such a recent endeavour is ``GEITje'' a model originally derived from the English-based Mistral 7B. Building on this fundamental work, the current research extends the capabilities of GEITje by supervised finetuning on newly created high-quality synthetic conversational datasets, along with an additional preference alignment procedure on a synthetic feedback dataset. Both the developed models and the created datasets are openly available.
\end{abstract}

\section{Introduction}

The presence of open-source language models has been marked by their successes in English. However, the linguistic reach of such technologies on non-English languages is often sidelined. As such, non-English speakers are forced to either rely on language technology in a foreign language, need to use closed-source or commercial systems, or they are restricted from leveraging the full potential of digital advancements altogether. In this paper, the focus lies specifically on Dutch, a language spoken by approximately 24 million people globally. The development of robust language technologies in Dutch not only democratizes access but also improves linguistic inclusivity, ensuring that all users can engage with technology in their native language.

The development of Dutch generative language models has lagged, with few initiatives aiming to bridge the gap left by English-centric models. This is unsurprising, since there is a significant cost involved in pretraining a state-of-the-art large language model (LLM). Yet, \citet{rijgersberg2023geitje} showed successfully that there is no need to pretrain a model from-scratch to achieve impressive model quality. Instead, he continue-pretrained an English-focused model \citep[Mistral 7B v0.1;][]{jiang2023mistral} on 10 billion tokens of Dutch data. In addition to the release of the model itself, his work importantly highlights the potential of continuous pretraining of existing models on Dutch. In addition to the base model, \citet{rijgersberg2023geitje} also introduce a chat variant of the model (GEITje Chat), which was finetuned using two synthetic datasets for supervised finetuning (SFT; also known as instruction tuning), automatically translated from originally English datasets with GPT 3.5, totalling around 20,000 conversations: No Robots NL (10,000 conversations)\footnote{\url{https://huggingface.co/datasets/Rijgersberg/no_robots_nl}} and Ultrachat 10k NL (a translated, random subset of the original English dataset).\footnote{\url{https://huggingface.co/datasets/Rijgersberg/ultrachat_10k_nl}} Using synthetic data, either translated or generated from-scratch with a different LLM, is common, even for English, because handcrafting conversations for large volumes of data is expensive and time-consuming -- although not impossible \citep[e.g.,][]{no_robots,DatabricksBlog2023DollyV2}.

The current paper delivers new high quality, high volume datasets (Section~\ref{sec:datasets}). Rather than translating original English datasets (both the prompt and the answer), two different approaches are considered to create new conversational datasets for SFT, both using GPT-4 (\texttt{gpt-4-1106-preview} via the Azure API) rather than GPT-3.5. For one dataset, only the prompt is translated but the answer is generated to avoid translationese and errors in translations. For another, large dataset an extensive approach is considered where an original English prompt is used as a seed, and a multi-turn conversation is generated by GPT-4 which chats with itself. On top of that, in that case the synthetic user responses are also modelled towards specific user groups to improve the diversity of the dataset. 

In addition to releasing improved datasets for SFT, this paper also contributes the first preference dataset of aligning LLMs for Dutch. Alignment is the process of ensuring that language models behave according to a given preference, for instance to avoid toxicity, to increase truthfulness, or simply to increase the output quality and conversational nature of a model. Different algorithms have been proposed for preference alignment. The most well-known example is Proximal Policy Optimization \citep{schulman2017proximal} as applied in the InstructGPT paper \citep{ouyang2022training}, where an additional reward model is created to rate and improve the output of the SFT model via reinforcement learning to maximize the estimated reward. A computationally more lenient approach was introduced with Direct Preference Optimization \citep[DPO;][]{rafailov2023direct}, which does not require a separate reward model but instead embeds the optimal policy in the model itself and adding a classification loss to determine which of the given answers to a given prompt are preferred. The dataset that we release is therefore compatible with such approaches that require a given input query and a ``rejected'' and a ``chosen'' answer. In the dataset, GPT-4 was always the chosen answer and the original GEITje chat always the rejected dataset\footnote{\url{https://huggingface.co/datasets/BramVanroy/ultra_feedback_dutch}}, but see the discussion in Section~\ref{sec:datasets} for improvements on this dataset.

These two types of datasets (SFT and preference) have enabled the creation of two models: GEITje 7B Ultra SFT and GEITje 7B Ultra. The former is a direct continuation of the original GEITje, finetuned on the supervised datasets to improve its conversational skills. The latter represents a further refinement, trained on the preference dataset to optimize the model's alignment with preference data. GEITje 7B Ultra stands as the first preference-optimized LLM for Dutch. ``Ultra'' is preferred to be used over the Ultra SFT version; the latter merely serving as an intermediate step to train Ultra. Both the datasets and models are publicly available.

\section{Datasets}\label{sec:datasets}

Two types of datasets are introduced: new, large datasets for supervised finetuning as well as novel datasets for preference tuning.

All datasets were filtered in the same manner after creation:

\begin{itemize}
    \item Language identification was added to all generated columns. All samples that were not marked as Dutch were removed.
    \item Samples that contain non-Latin characters were removed.
    \item Samples that have occurrences like `AI-assistent'' or ``AI-model'' were removed to avoid responses in the vain of ``As an AI model, ...''
    \item Samples that mention ChatGPT, GPT-3 or GPT-4, or ShareGPT were removed
    \item Samples that mention any kind knowledge cut off were removed
    \item Samples that contain apologies such as ``sorry'' or ``spijt me'' were removed
\end{itemize}

For all datasets, the Azure API was used to call on GPT4 \texttt{gpt-4-1106-preview} so the content filters may also have blocked some of the data generation queries in exceptional cases (for instance when asked about a topic pertaining to self-harm), which explains why the Dutch variants presented here contain fewer samples than the original English datasets.

\subsection{Supervised Finetuning}

\subsubsection{Ultra Chat 200k Dutch} 

Ultra Chat 200k Dutch is a recreation of the full UltraChat 200k dataset \citep{tunstall2023zephyr}, which itself is a filtered version of the original UltraChat dataset \citep{ding2023enhancing}. It is different from the subset that \citet{rijgersberg2023geitje} used in two ways: we use GPT-4 rather than GPT-3.5, and rather than translating the data we generate it fully from-scratch by using the original data as a seed. The GPT-4 API was prompted to converse with itself in multiple turns about a given topic. This topic was taken from the aforementioned UltraChat 200k dataset, i.e. the first English user message was provided as a seed and GPT-4 was prompted to converse about anything in relation to it, taking on alternating ``user'' and ``assistant'' roles, but in Dutch. Furthermore, to emphasise diversity and coverage, the model was instructed to take on a given user persona when providing a user response. The prompt to achieve this is given in Appendix~\ref{app:ultrachat-prompt}. Nine possible personas were created with a given description, which were randomly sampled alongside a given distribution (probability between brackets):

\begin{itemize}
    \item taalleerder (0.01): ``Deze persoon spreekt niet goed Nederlands en gebruikt geen moeilijke woorden of ingewikkelde zinsconstructies. Af en toe schrijft de persoon fouten, maar niet altijd.''\\
    EN: language learner: This person does not have a good command of Dutch and does not use difficult words or complicated sentence structures. Occasionally the person writes mistakes, but not always.
    \item direct (0.1): ``Een direct persoon die kortdadige taal hanteert. De gebruiker stelt specifieke, doelgerichte vragen in bondige en soms zelfs droge taal. De persoon verkiest een korte, duidelijke uitleg boven een lange, gedetailleerde uitleg.''\\
    EN: direct: A direct person who uses short -term language. The user asks specific, targeted questions in concise and sometimes even dry language. The person prefers a short, clear explanation over a long, detailed explanation.
    \item detailliefhebber (0.1): ``Een geduldig persoon die diepgaande vragen stelt en gedetailleerde antwoorden verwacht.''\\
    EN: details: A patient person who asks in-depth questions and expects detailed answers.
    \item kritisch (0.03): ``Een kritisch persoon die alles in vraag stelt en vaak moeilijk te overtuigen is.''\\
    EN: critical: A critical person who questions everything and is often difficult to convince.
    \item kind (0.01): ``Een jong persoon tussen 6 en 12 jaar oud die nog vele zaken niet kent en dus vragen stelt die voor ouderen misschien vanzelfsprekend zijn. Ook kan het zijn dat de persoon nog niet erg goed kan lezen en schrijven en dus zal de persoon zelf geen moeilijk taal gebruiken en soms om verduidelijking vragen.''\\
    EN: child: A young person between 6 and 12 years old who is not knowledgeable about many things yet and therefore asks questions that may be self-evident for others. It is also possible that the person cannot yet read and write very well and so the person will not use a difficult language and sometimes ask for clarifications.
    \item expert (0.15): ``Een ervaren expert die erg goed op de hoogte is van het onderwerp en dus ook diepgaande, bijna encyclopedische of academische, vragen stelt om wellicht een vak-specifiek probleem op te lossen.''\\
    EN: expert: An experienced expert who is very well aware of the subject and therefore also asks in-depth, almost encyclopedic or academic questions to perhaps solve a domain-specific problem.
    \item lachebek (0.01): ``Een persoon die graag lacht en grapjes maakt en in luchtige taal communiceert. De persoon gebruikt soms (maar niet altijd) smileys en andere emoticons om zijn/haar gevoelens te uiten. De persoon is voornamelijk geïnteresseerd in wonderbaarlijke en fantastische zaken en springt al eens van de hak op de tak.''\\
    EN: joker: A person who likes to laugh and makes jokes, and who communicates in informal language. The person sometimes uses smileys and other emoticons (but not always) to express his/her feelings. The person is mainly interested in wonderful and fantastic things and may jump from one topic to another.
    \item generalist (0.15): ``Een persoon die graag over veel verschillende onderwerpen praat en dus ook veel uiteenlopende vragen stelt. De persoon is niet erg geïnteresseerd in de details van een onderwerp, maar eerder in de grote lijnen.'' \\
    EN: generalist: A person who likes to talk about many different topics and therefore also asks many different questions. The person is not very interested in the details of a subject, but rather in the broad strokes of it.
    \item gemiddeld (0.44): ``Een gemiddelde, normale gebruiker die geen bijzonder eisen stelt of noden heeft maar simpelweg een behulpzame assistent verwacht.''\\
    EN: average: An average, typical user who does not have any specific demands but simply expects a helpful assistant.
\end{itemize}

By having a dataset that contains user messages that were written from different viewpoints (e.g. a child, or an expert), and high-quality GPT-4 answers to those queries, the ambition is that models trained with this type of data are more responsive to such a broad spectrum of users. Accessibility is a key concept for this dataset.

The dataset is focused around three main topics, just like the original UltraChat dataset \citep{ding2023enhancing}: questions about the world, going from technology, to art and entrepreneurship; creativity and writing, such as drafting emails or plays; and assistance with existing, given materials, e.g. rewriting, summarizing, and so on. It contains 192,598 conversations in the training set and 21,424 in the test set.

\subsubsection{No Robots Dutch} 

No Robots Dutch is an adaptation of the original No Robots dataset \citep{no_robots} to Dutch.\footnote{\url{https://huggingface.co/datasets/BramVanroy/no_robots_dutch}} The original dataset contains 10,000 manually created conversations in 10 categories: generation, open question answering, brainstorming, chat, rewriting, summarization, coding, classification, closed question answering, and answer extraction. Rather than translating the whole dataset to Dutch, like \citep{rijgersberg2023geitje}, only the user prompt was translated to Dutch with GPT-4. Then, this Dutch ``user prompt'' was answered by GPT-4 to avoid translation errors or translationese effects. After filtering, the training set contains 8181 samples and the test set 433. Some of the items in this dataset have a system message, which was also automatically translated.

\subsection{Preference Alignment}

\subsubsection{Ultra Feedback Dutch (Cleaned)}

Ultra Feedback Dutch\footnote{\url{https://huggingface.co/datasets/BramVanroy/ultra_feedback_dutch}} is a synthetic dataset specifically designed for the preference tuning of Dutch large language models. It contains 48,228 training and 5,359 test samples. The dataset is based on the original English Ultra Feedback dataset \citep{cui2023ultrafeedback} and is designed for use with models that have already undergone instruction tuning. The dataset comprises single-turn responses, generated by two distinct large language models, in response to a given prompt. One of the responses is ``chosen'' and the other ``rejected''. Preference algorithms such as Direct Preference Optimalization \citep{rafailov2023direct} will then push the models towards the preferred manner of responding.

To create Ultra Feedback Dutch, the first user message in the original English Ultra Feedback dataset was translated to Dutch with GPT4. This user message was then used as a starting point. Responses were then generated with GPT4 and GEITje-7B-chat\footnote{\url{https://huggingface.co/Rijgersberg/GEITje-7B-chat}} (GEITje). In the first version of the dataset, GPT4 was always chosen as the ``chosen'' best response and GEITje as the rejected response. GEITje 7B Ultra was trained on this initial dataset approach.

At a later stage, a cleaned version of Ultra Feedback Dutch was released.\footnote{\url{https://huggingface.co/datasets/BramVanroy/ultra_feedback_dutch_cleaned}} First of all, the responses by GEITje-7B-chat were replaced by those from GEITje-7B-Ultra. Secondly, GPT4 was not always selected as the best answer. Instead, the GEITje Ultra and GPT4 responses were rated by GPT4. GPT4 was used as LLM-as-a-judge. See for instance \citet{zheng2023judging}, who showed that GPT-4 can match controlled and crowdsourced human preferences. GPT4 rated all responses out of 5 on ``Dutch-ness'' (how grammatically correct and fluent is the response), Helpfulness (how helpful/relevant is the response), and Conciseness (how to-the-point is the response). The full prompt for scoring is given in Appendix~\ref{app:ultrafeedback-score-prompt}.

Two configurations of the dataset are given. The first one contains all original data, and the ``chosen'' response (GPT4 or GEITje 7B Ultra) is determined by the highest average score across the three criteria. In a smaller, high-quality subset only data where the average score of both responses is at least 4.0 is retained. Furthermore, if any of the models score less than 3.5 on any of the criteria, the sample is discarded. Finally, the absolute difference between the two models' average scores cannot be less than 0.25 or higher than 2.0. This ensure that the dataset contains ``competitive'' samples. The model with the highest average score is chosen as chosen, the other as rejected. In case of a tie, GPT4 wins.

\subsubsection{Orca DPO Pairs Dutch (Cleaned)}

Orca DPO Pairs Dutch\footnote{\url{https://huggingface.co/datasets/BramVanroy/orca_dpo_pairs_dutch}} is very similar to Ultra Feedback Dutch in how it was created. It was inspired by Intel's Orca DPO Pairs\footnote{\url{https://huggingface.co/datasets/Intel/orca_dpo_pairs}} which in turn is a subset of OpenOrca \citep{OpenOrca}. It contains around 10,000 single-turn conversations, where the English user message was translated with GPT4 to Dutch alongside a system message (if applicable). This Dutch system message and Dutch user message were then fed to GEITje 7B Chat and GPT4 to generate responses. In the first version, GPT4 was always selected as the ``chosen'' response.

This dataset was also later improved, although it was not scored and filtered like Ultra Feedback Dutch Cleaned.\footnote{\url{BramVanroy/orca_dpo_pairs_dutch_cleaned}} GEITje 7 Chat responses were replaced by GEITje 7B Ultra and duplicates were removed, leaving 9,437 items for training and 1052 for testing. GPT4 is still, naively, the ``chosen'' response.

The Dutch Orca dataset was not used for training GEITje 7B Ultra.

\section{Training Procedure}

(Part) of the datasets listed above were crucial for training GEITje 7B Ultra, which is discussed below. GEITje 7B Ultra's main goal was to be ``aligned'' with preference data. This was achieved through a two-step training process. The first step involved teaching the base model GEITje-7B \citep{rijgersberg2023geitje} how to follow instructions with supervised finetuning (SFT). This instruction model was then further aligned with preference data, leading to the intended model of this research: GEITje 7B Ultra.

\subsection{GEITje 7B Ultra SFT}

In the first step, GEITje 7B is ``taught'' to follow instructions. This version is simply called ``GEITje 7B Ultra SFT'', but its use is not recommended. It only serves as an intermediate step for creating the recommended GEITje 7B Ultra model.

The instruction datasets that were used are in part described in earlier sections and others (Alpaca Cleaned Dutch, Dolly 15K Dutch, Stack Overflow Chat Dutch) in prior work \citep{vanroy2023language}. The used datasets are summarized below. The percentages indicate their proportion in the total dataset, counted by the number of samples.

\begin{enumerate}
    \item ultrachat\_200k\_dutch\footnote{\url{https://huggingface.co/datasets/BramVanroy/ultrachat_200k_dutch}}: (gpt-4-turbo; multi-turn; generated): 85.42%
    \item no\_robots\_dutch\footnote{\url{https://huggingface.co/datasets/BramVanroy/no_robots_dutch}}: (gpt-4-turbo; prompt translate, answer generated; some items have system messages): 2.20%
    \item stackoverflow-chat-dutch\footnote{\url{https://huggingface.co/datasets/BramVanroy/stackoverflow-chat-dutch}}: (gpt-3.5-turbo; multi-turn; code; translated; only 50\% used): 8.38%
    \item alpaca-cleaned-dutch\footnote{\url{https://huggingface.co/datasets/BramVanroy/alpaca-cleaned-dutch}}: (gpt-3.5-turbo; translated): 2.62%
    \item dolly-15k-dutch\footnote{\url{https://huggingface.co/datasets/BramVanroy/dolly-15k-dutch}}: (gpt-3.5-turbo; translated): 1.39%
\end{enumerate}

The full training set consists of 240,527,565 tokens (calculated prior to applying a chat template). The test set accounts for 26,397,086 tokens, which is around 10.97\% of the training set.

The model was trained in bfloat16 with flash attention 2 and a context length of 8192 on two nodes of four A100 80GB each for around 2.5 hours on the Flemish SuperComputer.\footnote{\url{https://www.vscentrum.be/compute}} The Zephyr chat template was used. Full hyperparameters are given in Appendix~\ref{app:training-GEITje-7B-ultra-sft}.

\subsection{GEITje 7B Ultra}

The second step to create GEITje 7B Ultra involved the application of the Direct Preference Optimisation algorithm \citep{rafailov2023direct} to the GEITje-7B-ultra-sft model. The goal is to align the model more closely with user preferences. To do so, the model was trained on the aforementioned Ultra Feedback Dutch dataset, amounting to 56,137,090 tokens (combination of prompt, rejected and chosen columns) and 6,178,969 tokens respectively for training and testing.

Similar to GEITje 7B Ultra SFT, the final DPO model was trained in bfloat16 with flash attention 2 and a context length of 8192 on two nodes of four A100 80GB each for around 11 hours on the Flemish SuperComputer. Full hyperparameters are given in Appendix~\ref{app:training-GEITje-7B-ultra}.

Interestingly, the beta parameter in the DPO algorithm seemed hard to tune. A lower beta value allows the model to deviate more from the base model, emphasizing the preference data more strongly. Conversely, a higher beta value constrains the model to stay closer to the base model, making smaller adjustments to align with the preference data. The recommended value of $0.01$, as described in \citet{tunstall2023zephyr}, did not work for this model and dataset combination, leading to repetitions and hallucinations. After a hyperparameter search, the best value of $beta=0.1$ was selected.

\section{Results}

To evaluate GEITje 7B Ultra, we refer to the results provided by the ScandEval leaderboard.\footnote{\url{https://scandeval.com/dutch-nlg/}} For details and exact implementation, see \cite{nielsen2023scandeval,nielsen2024encoder}, and notably their dataset configurations on GitHub, which includes the subset, exact prompts and number of few-shot examples used.\footnote{\url{https://github.com/ScandEval/ScandEval/blob/8766d2aef43d86eda4f6d86dc7ae2fea0142947e/src/scandeval/dataset_configs.py}} ScandEval is a leaderboard and Python package to evaluate LLMs, focused on Scandinavian and Germanic languages.

A summary of the tasks listed here:

\begin{itemize}
    \item CoNLL NL: entity recognition dataset, from the CoNLL 2002 dataset \citep{tksintro2002conll}
    \item Dutch Social: a dataset of Dutch tweets, intended for sentiment analysis \citep{aakash2020dutchsocial}
    \item ScaLA NL: the Dutch portion of a linguistic acceptability benchmark \citep{nielsen2023scandeval}
    \item SQuAD NL: a Dutch machine-translated version of the English questions-and-answers dataset SQuAD v2. \textbf{Note}: multiple translated versions of SQuAD exist. In ScandEval, the translation from Yeb Havinga are used.\footnote{\url{https://huggingface.co/datasets/yhavinga/squad_v2_dutch}}
    \item WikiLingua NL: the Dutch portion of the WikiLingua summarization dataset \citep{ladhak-etal-2020-wikilingua}
    \item MMLU NL: a Dutch machine-translated version by \cite{dac2023okapi} of the English multiple choice questions (MCQ) dataset Massive Multitask Language Understanding \citep{hendrycks2021measuring}
    \item HellaSwag NL: a Dutch machine-translated version by \cite{dac2023okapi} of the English common-sense reasoning dataset HellaSwag \citep{hendrycks2021measuring}
\end{itemize}

\begin{landscape}
\begin{table}[!h]
\centering
\begin{tabular}{l|r|r|r|r|r|r|r||r}
\textbf{model\_id} & \multicolumn{1}{l|}{\textbf{conll\_nl}} & \multicolumn{1}{l|}{\textbf{dutch\_social}} & \multicolumn{1}{l|}{\textbf{scala\_nl}} & \multicolumn{1}{l|}{\textbf{squad\_nl}} & \multicolumn{1}{l|}{\textbf{wiki\_lingua\_nl}} & \multicolumn{1}{l|}{\textbf{mmlu\_nl}} & \multicolumn{1}{l||}{\textbf{hellaswag\_nl}} & \multicolumn{1}{l}{\textbf{average}} \\ \hline
gpt-4-1106-preview & \textbf{66.44} & \textbf{14.22} & \textbf{72.30} & 57.81 & 67.13 & \textbf{70.04} & \textbf{88.29} & \textbf{62.32} \\ \hline
mistralai/Mistral-7B-v0.1 & 58.15 & 7.94 & 25.41 & \textbf{62.56} & 64.24 & 35.49 & 19.88 & 39.10 \\ \hline
BramVanroy/GEITje-7B-ultra & 42.20 & 12.78 & 18.23 & 53.41 & \textbf{68.30} & 26.92 & 25.72 & 35.37 \\ \hline
Rijgersberg/GEITje-7B & 47.53 & 4.36 & 30.67 & 56.55 & 67.58 & 28.12 & 11.70 & 35.22 \\ \hline
Rijgersberg/GEITje-7B-chat-v2 & 42.12 & 11.06 & 19.71 & 59.19 & 65.55 & 27.71 & 18.03 & 34.77 \\ \hline
Rijgersberg/GEITje-7B-chat & 50.69 & 8.16 & 20.45 & 54.48 & 66.92 & 24.89 & 9.84 & 33.63 \\ \hline
BramVanroy/GEITje-7B-ultra-sft & 39.41 & 7.00 & 16.10 & 53.02 & 67.74 & 25.80 & 12.49 & 31.65
\end{tabular}

    \caption{Benchmark results from ScandEval for GEITje models as well as the base model Mistral 7B and the GPT4 model that was used to create much of the data discussed here.}
\end{table}
\end{landscape}

\section{Discussion}

The goal of creating GEITje 7B Ultra was to align it with preference data in order to improve its appropriateness as an assistant. In other words, the alignment step was mostly directed towards its style and manner of communicating and not directly intended to improve its benchmark performance. An additional requirement was that benchmark results should at least be comparable to the original GEITje 7B Chat.

Looking more closely at the results we can see that the average results of GPT4 are still far ahead of all GEITje models. Interestingly, this is especially the case for the MMLU knowledge questions and commonsense reasoning but also for the linguistic acceptability test. A note here is that the training data of GPT4 is not known, so for these benchmarks it is not clear whether GPT4 was perhaps trained on the test data. Comparing GEITje 7B Ultra with GEITje 7B Chat v2, another interesting tendency appears: both models perform very similarly in most tasks, but differences are larger in question answering (Chat v2 better) and commonsense reasoning (Ultra better). Another observation is that the SFT version prior to Ultra performs worse than all other models whereas the base GEITje-7B version performs better than all GEITje versions except for Ultra. Useful future work would be to provide ablations on the datasets, comparing the datasets that were used in Chat v2 and Ultra SFT. Importantly, Ultra performs on par with Chat v2, as intended.

An important note here, that was made earlier in \citet{vanroy2023language}, is that these benchmarks should be interpreted correctly. They do not measure usefulness in the Dutch language. High scores do not necessarily correlate with high fluency of Dutch. This is evident from Mistral 7B's high score, which in practical use as an assistant is far less usable than any of the GEITje models when it comes to language proficiency. In these benchmarks, the models are prompted to answer multiple-choice questions or to extract the right words from a given text. That is a useful task but does make any assurances when it comes to usefulness as an assistant in the Dutch language, for instance. These benchmarks do not query the models to write running text. While these benchmarks seem to measure language understanding, they do not account for a model's fluent Dutch generative capabilities.

In this paper the creation of the relevant datasets as well as the training of the models has been described. Comparative results are provided thanks to ScandEval, indicating that indeed this iteration of GEITje maintains its prior abilities while being more fluent in its capacity as an assistant. All outcomes such as models, datasets and training code, are publicly available at \url{https://huggingface.co/collections/BramVanroy/geitje-7b-ultra-65c1ee010ad80fd1f6a8f208}.

\section*{Acknowledgments}

Thanks to Michiel Buisman of UWV for reaching out and making the creation of the datasets possible with access to Azure's API to query \texttt{gpt-4-1106-preview}. Another thank you to the Flemish Supercomputer (Vlaams Supercomputer centrum; \url{https://www.vscentrum.be/compute}) for access to the hardware to train the models.

Thanks to David Berenstein (Argilla) for providing an initial prompt for rating the responses, ultimately leading to the Ultra Feedback Dutch Cleaned dataset. Particular thanks to Edwin Rijgersberg for always being open to discuss LLMs and data for Dutch and developing GEITje 7B.

\bibliography{references}
\bibliographystyle{colm2024_conference}

\appendix
\section{UltraChat Creation: GPT-4 Prompt}\label{app:ultrachat-prompt}

As described on the dataset page.\footnote{\url{https://huggingface.co/datasets/BramVanroy/ultrachat_200k_dutch\#dataset-creation}}

\begin{lstlisting}[breaklines=true]
# Simulatie van Interactie Tussen een Gebruiker en een AI-assistent

Je simuleert een interactie tussen een gebruiker met een gegeven 'Persona' en een AI-assistent. De interactie wordt gestart op basis van een gegeven 'Startvraag'.

## Persona van Gebruiker

De gebruiker krijgt een specifieke 'Persona' toegewezen, die diens manier van communiceren en de persoonlijkheid omschrijft. Alles dat de gebruiker zegt moet dus in lijn zijn met de karaktereigenschappen en communicatiestijl van de toegewezen Persona. De AI-assistent gedraagt zich als een behulpzame assistent en moet de vragen van de gebruiker objectief, en zo goed en eerlijk mogelijk beantwoorden en de instructies juist volgen.

## Startvraag

Je krijgt een 'Startvraag' in het Engels mee als startpunt van de interactie. Dat kan een vraag of instructie zijn. Als eerste stap moet je deze startvraag vertalen naar het Nederlands en volledig aanpassen aan het taalgebruik en persona van de gebruiker zodat de gebruiker met deze aangepaste vraag of instructie het gesprek kan beginnen. Zorg ervoor dat ALLE inhoud van de oorspronkelijk vraag behouden blijft maar pas waar nodig de schrijfstijl grondig aan.

## Beurten

Na de startvraag antwoordt de assistent. Afhankelijk van de persona kan de gebruiker daarna vragen om meer details, gerelateerde informatie, het antwoord in vraag stellen, of de instructies verder verfijnen. Dat gebeurt in verschillende op elkaar voortbouwende interacties zoals in een echt gesprek. Het gesprek neemt tussen de 5 en 12 beurten van zowel de gebruiker als de assisent in beslag. Gebruikers met Persona's die meer vragen stellen, zullen dus meer beurten nodig hebben.

## Taalgebruik

De vragen, instructies en antwoorden moeten in het Standaardnederlands geschreven zijn tenzij anders aangegeven in de Persona van de gebruiker. De taal is verzorgd en bevat geen regionale variatie zodat het over het hele taalgebied (waaronder Nederland en Vlaanderen) zonder problemen begrepen kan worden.

## Input en Output Formaat

Als input krijg je een 'Persona' van de gebruiker en een 'Startvraag' of instructie in het Engels. Voorbeeld input:

```
<persona>
[Beschrijving van de Persona van de gebruiker]
</persona>
<startvraag>
[Een korte of lange vraag of instructie in het Engels die eerst vertaald moet worden en dan aangepast moet worden aan de persona]
</startvraag>
```

De output moet simpel gestructureerd zijn zodat je voor de gebruiker en assistent respectievelijk de gebruikersvraag of -instructie en het antwoord van de assistent geeft. 

Voorbeeld output:

```
gebruiker: [Vertaling en aanpassing van de Startvraag aan de persona in passend taalgebruik]
assistent: [antwoord op de vorige gebruikersvraag of -instructie]

gebruiker: [vervolgvraag-1]
assistent: [antwoord op de vorige vervolgvraag-1]

gebruiker: [vervolgvraag-2]
assistent: [antwoord op de vorige vervolgvraag-2]
```

---

<persona>
{persona}
</persona>
<startvraag>
{subject}
</startvraag>

\end{lstlisting}

\section{Ultra Feedback Dutch Cleaned Creation: GPT-4 Rating Prompt}\label{app:ultrafeedback-score-prompt}

As described on the dataset page.\footnote{\url{https://huggingface.co/datasets/BramVanroy/ultra_feedback_dutch_cleaned\#user-prompt}}

For every model we query GPT4 multiple times, once for each criterion. We investigated three criteria: Dutch-ness (how good is the model's Dutch output), Helpfulness (how relevant is the model's reply), and Conciseness (how to-the-point is the model).

Below you find the template and criteria. \texttt{criterion\_options} is a formatted list of the given options for a given criterion according to \texttt{opt\_template} for each option.

\begin{lstlisting}[breaklines=true]
template = """Het volgende is een instructie geschreven door een mens (`Instructie:`), en een reactie op de instructie geschreven door een AI-model (`Reactie:`). Beoordeel de kwaliteit van de reactie van het AI-model, rekening houdend met de gegeven opties (`Opties:`).

Instructie:
{prompt}

---

Reactie:
{response}

---

Criteria: {criterion_question}

Opties:
{criterion_options}

---

Je antwoord moet in het volgende formaat zijn:

<rating>[{{min_score}}-{{max_score}}]</rating>

bijvoorbeeld:

<rating>3</rating>

---

Beoordeel nu alsjeblieft de `Reactie:` met een rating op basis van de `Opties:`. Geef geen extra uitleg."""

opt_template = """\
- {score}: {beschrijving}\
"""


criteria = {
  "dutchness": {
    "criterion_question": "Is de reactie in vlot en gramaticaal correct Nederlands geschreven? Negeer code-fragmenten in je analyse en richt je enkel op de doorlopende tekst. Leenwoorden uit andere talen mogen gebruikt worden als dat gewoonlijk is in het domein (bv. bij software). Een hogere score duidt op beter Nederlands taalgebruik.",
    "criterion_options": {
      1: "De reactie is onleesbaar, bevat veel grammaticale fouten, of is in slecht Nederlands geschreven.",
      2: "De reactie is moeilijk te begrijpen of bevat veel grammaticale fouten.",
      3: "De reactie is begrijpelijk maar bevat enkele grammaticale fouten.",
      4: "De reactie is goed geschreven en bevat weinig grammaticale fouten.",
      5: "De reactie is uitstekend geschreven, vlot leesbaar en bevat geen grammaticale fouten.",
    },
  },
  "helpfulness": {
    "criterion_question": "Is de reactie relevant en behulpzaam? Beantwoordt het model de instructie goed? Een hogere score duidt op een relevantere en behulpzamere reactie.",
    "criterion_options": {
      1: "De reactie is helemaal niet relevant of heeft aanzienlijke afwijkingen.",
      2: "De reactie is slechts enigszins relevant maar is niet concreet.",
      3: "De reactie is min of meer relevant en geeft een relevant antwoord.",
      4: "De reactie is grotendeels relevant en lijkt zeer nuttig.",
      5: "De reactie biedt briljante ideeen die de taak nauwkeurig aanpakken.",
    },
  },
  "conciseness": {
    "criterion_question": "Is de reactie beknopt en ter zake, zonder onnodige herhaling of uitweiding? Een hogere score duidt op een beknoptere, duidelijkere reactie.",
    "criterion_options": {
      1: "De reactie bevat overmatige herhaling of onnodige uitweiding.",
      2: "De reactie is nogal omslachtig.",
      3: "De reactie is redelijk beknopt met minimaal onnodige inhoud.",
      4: "De reactie is beknopt en ter zake, met minimaal onnodige inhoud.",
      5: "De reactie is uitzonderlijk positief beknopt, verstrekt informatie efficient.",
    },
  },
}

\end{lstlisting}

\section{Training Details: GEITje-7B-ultra-sft}\label{app:training-GEITje-7B-ultra-sft}

Training config, to be used with the alignment handbook \citep{Tunstall_The_Alignment_Handbook}.\footnote{\url{https://github.com/huggingface/alignment-handbook}} Also provided in the model card.\footnote{\url{https://huggingface.co/BramVanroy/GEITje-7B-ultra-sft\#training-procedure}}

\begin{lstlisting}
# Model arguments
model_name_or_path: Rijgersberg/GEITje-7B
model_revision: main
torch_dtype: bfloat16
use_flash_attention_2: true

# Data training arguments
# Zephyr chat template
chat_template: "{% for message in messages %}\n
{% if message['role'] == 'user' %}\n
{{ '<|user|>\n' + message['content'] + eos_token }}\n
{% elif message['role'] == 'system' %}\n
{{ '<|system|>\n' + message['content'] + eos_token }}\n
{% elif message['role'] == 'assistant' %}\n
{{ '<|assistant|>\n'  + message['content'] + eos_token }}\n
{% endif %}\n
{% if loop.last and add_generation_prompt %}\n
{{ '<|assistant|>' }}\n
{% endif %}\n
{% endfor %}"
dataset_mixer:
  BramVanroy/ultrachat_200k_dutch: 1.0
  BramVanroy/stackoverflow-chat-dutch: 0.5
  BramVanroy/alpaca-cleaned-dutch: 1.0
  BramVanroy/dolly-15k-dutch: 1.0
  BramVanroy/no_robots_dutch: 1.0
dataset_splits:
- train_sft
- test_sft
preprocessing_num_workers: 8

# SFT trainer config
bf16: true
do_eval: true
evaluation_strategy: epoch
gradient_accumulation_steps: 1
gradient_checkpointing: true
gradient_checkpointing_kwargs:
  use_reentrant: False
hub_model_id: GEITje-ultra-sft
hub_strategy: every_save
learning_rate: 2.0e-05
log_level: info
logging_steps: 5
logging_strategy: steps
lr_scheduler_type: cosine
max_seq_length: 8192
max_steps: -1
num_train_epochs: 1
output_dir: data/GEITje-ultra-sft
overwrite_output_dir: true
per_device_eval_batch_size: 8
per_device_train_batch_size: 16
push_to_hub: true
remove_unused_columns: true
report_to:
- wandb
save_strategy: "steps"
save_steps: 100
save_total_limit: 1
seed: 42
warmup_ratio: 0.1
\end{lstlisting}

\section{Training Details: GEITje-7B-ultra}\label{app:training-GEITje-7B-ultra}

Training config, to be used with the alignment handbook \citep{Tunstall_The_Alignment_Handbook}.\footnote{\url{https://github.com/huggingface/alignment-handbook}} Also provided in the model card.\footnote{\url{https://huggingface.co/BramVanroy/GEITje-7B-ultra\#training-procedure}}

\begin{lstlisting}
# Model arguments
model_name_or_path: BramVanroy/GEITje-7B-ultra-sft
model_revision: main
torch_dtype: bfloat16
use_flash_attention_2: true

# Data training arguments
# For definitions, see: src/h4/training/config.py
dataset_mixer:
  BramVanroy/ultra_feedback_dutch: 1.0
dataset_splits:
- train_prefs
- test_prefs
preprocessing_num_workers: 8

# DPOTrainer arguments
bf16: true
beta: 0.1
do_eval: true
evaluation_strategy: steps
eval_steps: 100
gradient_accumulation_steps: 4
gradient_checkpointing: true
gradient_checkpointing_kwargs:
  use_reentrant: False
hub_model_id: BramVanroy/GEITje-ultra
learning_rate: 5.0e-7
log_level: info
logging_steps: 10
lr_scheduler_type: cosine
max_length: 2048
max_prompt_length: 1536
num_train_epochs: 1
optim: adamw_torch
output_dir: data/GEITje-ultra
per_device_train_batch_size: 4
per_device_eval_batch_size: 4
push_to_hub: true
save_strategy: "steps"
save_steps: 100
save_total_limit: 3
seed: 42
warmup_ratio: 0.1
\end{lstlisting}

\end{document}